\documentclass[10pt,twocolumn,letterpaper]{article}

\usepackage[pagenumbers]{iccv} 

\newcommand{\name}{DASH\xspace}

\definecolor{iccvblue}{rgb}{0.21,0.49,0.74}
\usepackage[pagebackref,breaklinks,colorlinks,allcolors=iccvblue]{hyperref}
\usepackage{multirow,threeparttable,algorithm,algpseudocode}
\usepackage{colortbl}  
\usepackage{xcolor}
\usepackage{array}   
\usepackage{marvosym}
\renewcommand{\thefootnote}{\fnsymbol{footnote}}
\usepackage{caption}
\usepackage[accsupp]{axessibility}  
\usepackage{ragged2e}

\definecolor{tablebest}{rgb}{1.0, 0.6, 0.6}
\definecolor{tablesecond}{rgb}{0.98, 0.78, 0.57}
\definecolor{tablethird}{rgb}{1.0, 1.0, 0.56}

\def\paperID{1878} 
\def\confName{ICCV}
\def\confYear{2025}

\title{DASH: 4D Hash Encoding with Self-Supervised Decomposition \\ for Real-Time Dynamic Scene Rendering }

\author{
    Jie Chen\textsuperscript{1}\and
    Zhangchi Hu\textsuperscript{1} \and
    Peixi Wu\textsuperscript{1} \and
    Huyue Zhu\textsuperscript{1} \and
    Hebei Li\textsuperscript{1,\dag} \and
    Xiaoyan Sun\textsuperscript{1,2,\dag} \and
\textsuperscript{1}University of Science and Technology of China \and
\textsuperscript{2}Institute of Artificial Intelligence, Hefei Comprehensive National Science Center\\
{\tt\small \{chenjie02, lihebei\}@mail.ustc.edu.cn, 
\{sunxiaoyan\}@ustc.edu.cn
}
}

\begin{document}

\twocolumn[{%
\renewcommand\twocolumn[1][]{#1}%
\maketitle

\vspace{-10pt}
\begin{center}
    \centering
    \captionsetup{type=figure}
    \includegraphics[width=1.0\textwidth]{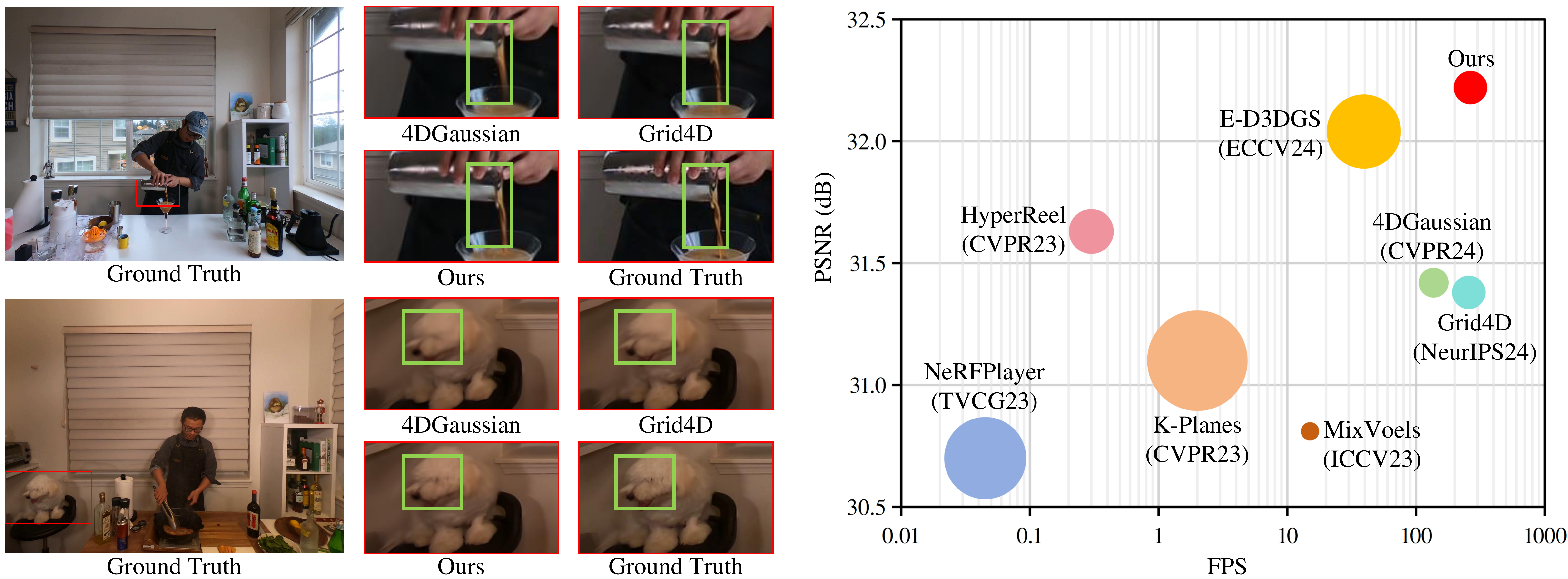}
    \captionof{figure}{Our method achieves real-time rendering for dynamic scenes with high rendering quality. In real-world datasets with intricate details, our method outperforms 4DGaussian~\cite{4dgaussian}, and Grid4D~\cite{grid4d} in terms of rendering quality and time performance. The right figure is tested on Neural 3D Video~\cite{dynerf} dataset, where the radius of the dot corresponds to the training time and and the horizontal axis is plotted on a logarithmic scale.}
\label{fig:teaser}
\end{center}%
}]

\newcommand\blfootnote[1]{%
  \begingroup
  \renewcommand\thefootnote{}\footnote{#1}%
  \addtocounter{footnote}{-1}%
  \endgroup
}
\blfootnote{\hspace{-1em} \textsuperscript{\dag}Corresponding authors}

\begin{abstract}
Dynamic scene reconstruction is a long-term challenge in 3D vision. Existing plane-based methods in dynamic Gaussian splatting suffer from an unsuitable low-rank assumption, causing feature overlap and poor rendering quality. Although 4D hash encoding provides an explicit representation without low-rank constraints, directly applying it to the entire dynamic scene leads to substantial hash collisions and redundancy. To address these challenges, we present \name, a real-time dynamic scene rendering framework that employs 4D hash encoding coupled with self-supervised decomposition.
Our approach begins with a self-supervised decomposition mechanism that separates dynamic and static components without manual annotations or precomputed masks. Next, we introduce a multiresolution 4D hash encoder for dynamic elements, providing an explicit representation that avoids the low-rank assumption. Finally, we present a spatio-temporal smoothness regularization strategy to mitigate unstable deformation artifacts.
Experiments on real-world datasets demonstrate that \name achieves state-of-the-art dynamic rendering performance, exhibiting enhanced visual quality at real-time speeds of \textbf{264} FPS on a single 4090 GPU. Code: \url{https://github.com/chenj02/DASH}.
\end{abstract}

\section{Introduction}
\label{sec:intro}
Novel view synthesis (NVS) is a critical 3D vision task, supporting VR, AR, and film production. It generates photorealistic scene renderings from arbitrary viewpoints/timestamps via 3D reconstruction from 2D images. While static scene modeling has high fidelity, dynamic reconstruction remains challenging due to complex nonlinear deformations and limited dense multi-view temporal data.

Traditional methods for dynamic scene reconstruction typically employed Neural Radiance Fields (NeRF)~\cite{nerf} combined with deformation fields. However, MLP-based deformation networks~\cite{d-nerf, fsdnerf, banmo, dynamic-nerf, dynibar, hypernerf, ndf, nerf-ds, dynerf, nerflow, hyperreel} often suffer from slow convergence and rendering artifacts.
Structured explicit representations, including plane-based~\cite{eg3d} and hash-grid~\cite{instant-ngp} methods, better preserves local geometry and accelerates training~\cite{tensor4d,k-planes,hexplane,4k4d,tineuvox,masked-spacetime-hashing,forwardflowdnerf,factorized-motion,nerfplayer}.
Recently, 3D Gaussian Splatting (3DGS)~\cite{gaussian-splatting} has advanced static scene rendering through real-time Gaussian representations. This method has been extended to dynamic scenes by incorporating timestamp-conditioned deformation fields~\cite{4dgaussian,deformable-3d-gaussians,dn4dgs,motiongs,spacetime-gaussian,grid4d,ed3dgs,3dgstream}.

\begin{figure}[tbp]
    \centering
    \includegraphics[width=8.3cm]{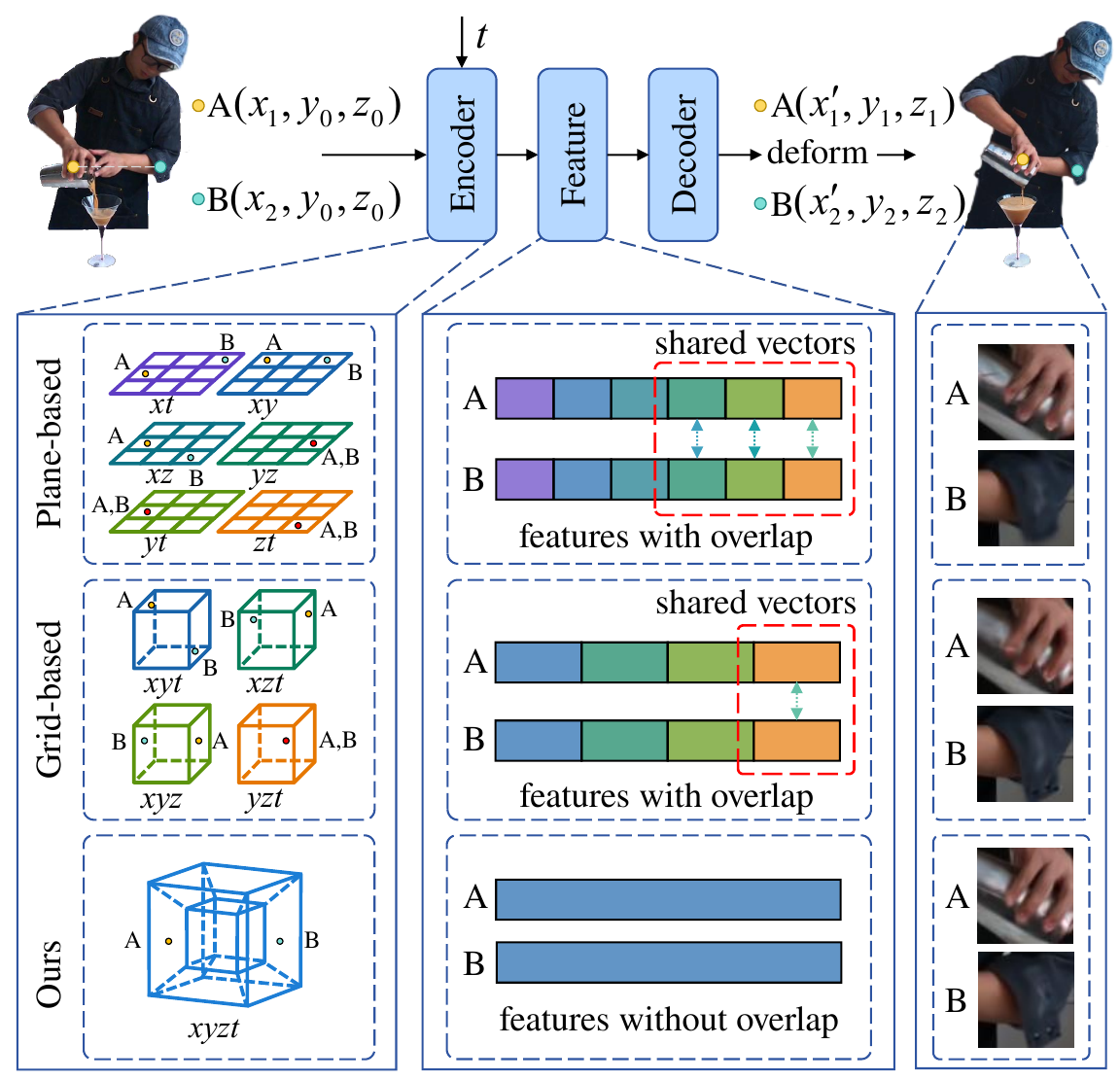}
    \caption{Comparison of feature encoding. Compared to the plane-based method~\cite{4dgaussian} and grid-based method~\cite{grid4d} which concatenate low-rank features into 4D features, our method eliminates the overlap in the features when encoding points like A and B with heavily overlapping coordinates.}
    \label{fig:feature_overlap}
\end{figure}

In dynamic novel view synthesis, both implicit and explicit neural networks have been employed to deform Gaussians. However, implicit MLP networks tend to over-smooth details, while explicit representations face other challenges. For example, plane-based methods such as 4DGaussian~\cite{4dgaussian} decompose the 4D spatiotemporal encoding into six 2D planes. This approach relies on a low-rank assumption that deformation features share significant commonality and can be factorized into a compact representation~\cite{tensor4d,k-planes,hexplane,4dgaussian,grid4d}. However, this assumption is unsuitable, often resulting in feature overlap and reduced accuracy of deformation predictions, as shown in \cref{fig:feature_overlap}. Although the grid-based method Grid4D~\cite{grid4d} avoids the low-rank assumption by decomposing the 4D encoding into four 3D hash encodings, it still suffers from feature overlap that limits its performance. Instead of decomposing the 4D spatiotemporal features, 4D hash encoding provides a promising explicit representation without low-rank assumption, and eliminates the overlap in the features. However, directly applying a 4D hash encoding to a dynamic scene would require a larger hash table size due to the more hash collisions caused by the additional time dimension~\cite{masked-spacetime-hashing}.

To address these problems, we propose \name, a real-time dynamic scene rendering framework that employs 4D hash encoding coupled with self-supervised decomposition.
Our key observation is that most scenes contain large static regions, which do not need to be stored in the 4D hash table. Based on this, we introduce a self-supervised dynamic-static decomposition method. This method leverages the inherent behavioral difference between static and dynamic elements during optimization, without requiring manual annotations or precomputed masks. Subsequently, static regions are reconstructed using standard Gaussian splatting~\cite{gaussian-splatting}, while the deformation fields of dynamic regions are modeled using our novel 4D hash encoder without relying on the low-rank assumption. Finally, to resolve the inherent lack of smoothness in explicit representations~\cite{grid4d}, we propose a spatio-temporal smoothness regularization that reduces unstable deformation artifacts and improves overall rendering fidelity.
Experimental results demonstrate that \name significantly outperforms other methods in both visual quality and rendering speed.

In summary, our main contributions are as follows:
\begin{itemize}
    \item We introduce a novel self-supervised method to decompose dynamic scenes into static and dynamic components without manual annotations or precomputed masks.
    \item We propose a novel explicit representation for dynamic scene rendering that leverages 4D hash encoding without relying on the low-rank assumption, and we apply a smooth regularization that effectively suppresses erratic deformation predictions.
    \item Experiments show that \name achieves state-of-the-art performance, with real-time rendering up to 264 FPS at a resolution of $1352\times1014$.
\end{itemize}

\section{Related Works}
\label{sec:related}

\subsection{NeRF-Based Dynamic Scene Rendering}
Novel view synthesis is a fundamental yet challenging task in 3D reconstruction. NeRF~\cite{nerf} pioneered implicit static scene representation using MLPs, and subsequent works~\cite{d-nerf,fsdnerf,banmo,dynamic-nerf,dynibar,hypernerf,ndf,nerf-ds,dynerf,nerflow,li2021neural,ndr,nr-nerf,rodynrf,s-nerf,saff,4dregsdf,star} extended it to dynamic scenes by incorporating canonical 3D grids and deformation fields.
D-NeRF~\cite{d-nerf} introduces implicit deformation fields on static models, providing an effective framework for dynamic scene rendering. HyperNeRF~\cite{hypernerf} models object topology deformations using higher-dimensional representations, while DyNeRF~\cite{dynerf} leverages time-conditioned NeRF for 4D scene modeling.
However, MLP-based implicit representations suffer from inherent over-smoothing and high computational costs due to dense ray sampling and volume rendering.

To address these limitations, explicit representations such as triplanes~\cite{eg3d} and hash encoding~\cite{instant-ngp} improve NeRF by enhancing visual quality and training efficiency. Several plane-based explicit representations, including K-Planes~\cite{k-planes}, HexPlane~\cite{hexplane}, Tensor4D~\cite{tensor4d}, and 4K4D~\cite{4k4d}, decompose 4D inputs into six 2D feature planes to facilitate dynamic scene rendering.
Additionally, hash encoding and 3D grid-based explicit representations assist MLPs in predicting deformations with improved speed and accuracy~\cite{tineuvox, masked-spacetime-hashing, forwardflowdnerf, mixvoels, nerfplayer}.
Despite these advancements, existing NeRF-based methods still struggle to meet real-time performance requirements.

\subsection{Gaussian-Based Dynamic Scene Rendering}
Recently, 3D Gaussian Splatting (3DGS)~\cite{gaussian-splatting} introduced anisotropic 3D Gaussians, initialized from structure-from-motion (SfM), to represent 3D scenes. In this approach, each 3D Gaussian is optimized volumetrically and projected to 2D for rasterization. By integrating pixel colors using $\alpha$-blending, 3DGS achieves high-quality rendering with fine details while maintaining real-time performance.

Subsequent works~\cite{4dgaussian,deformable-3d-gaussians,dn4dgs,dynamic3dgs,motiongs,spacetime-gaussian,grid4d,sc-gs,4dgs,ed3dgs,gags,3dgstream,cogs,swings,sarogs,swift4d} extended 3DGS to dynamic scenes by introducing deformation mechanisms for Gaussian attributes.While full MLP-based deformation fields offer high rendering quality~\cite{deformable-3d-gaussians}, they suffer from over-smoothing, limiting their ability to reconstruct fine details and represent complex scene dynamics. Furthermore, 4DGaussian~\cite{4dgaussian} employs a plane-based explicit representation that decomposes 4D space-time encoding into six 2D planes. It relies on an unsuitable low-rank assumption that deformation features share significant commonality and can be factorized into a compact representation~\cite{tensor4d,k-planes,hexplane,4dgaussian,grid4d}. Grid4D~\cite{grid4d} proposes an alternative approach by decomposing 4D encoding into one spatial 3D hash grid and three temporal 3D hash grids without the low-rank assumption. However, the resulting features still suffer from overlapping issues, reducing their discriminative power for deformation prediction.
Our work mainly focuses on mitigating feature overlap in both plane-based and hash grid explicit representations, improving the rendering quality of Gaussian-based models for dynamic scene reconstruction.

\section{Preliminary: 3D Gaussian Splatting}
\label{sec:pre}

3D Gaussian Splatting~\cite{gaussian-splatting} represents the scene using a set of 3D Gaussians. Each Gaussian is defined by a opacity $\sigma \in [0,1]$, a center position $\mathbf{\mu} \in \mathbb{R}^{3\times1}$, and a covariance matrix $\mathbf{\Sigma} \in \mathbb{R}^{3\times3}$.
For any point $\mathbf{p}=(x,y,z) \in \mathbb{R}^{3\times1}$ in 3D space, its contribution from a Gaussian is given by:
\begin{equation}
    G(\mathbf{p}) = \sigma \cdot e^{ -\frac{1}{2} (\mathbf{p}-\mathbf{\mu})^{\top} \mathbf{\Sigma}^{-1} (\mathbf{p}-\mathbf{\mu}) }.
    \label{eq:3dgs}
\end{equation}
Since the covariance matrix is positive and semi-definite, it can be decomposed into a scale $\mathbf{S}$ and a rotation $\mathbf{R}$ as $\mathbf{\Sigma} = \mathbf{R}\mathbf{S}\mathbf{S}^T\mathbf{R}^T$.
For novel view synthesis, the covariance matrix in camera coordinates is $\mathbf{\Sigma}^{\prime} = \mathbf{J}\mathbf{W}\Sigma \mathbf{W}^T\mathbf{J}^T$, where $\mathbf{W}$ is the view transformation matrix mapping world coordinates to camera space, and $\mathbf{J}$ represents the Jacobian matrix of the local affine approximation for perspective projection.

To render each pixel, spherical harmonics (SH) are employed to represent the color $c$ of 3D Gaussians. The color and the opacity of each 3D Gaussian are computed using \cref{eq:3dgs}. The final rendered color $C$ is obtained through $\alpha$-blending of the $N$ ordered 3D Gaussians overlapping the pixel as:
\begin{equation}
    C = \sum_{i}^N c_i \alpha_i \prod_{j=1}^{i-1} (1-\alpha_j),
    \label{eq:3dgs_color}
\end{equation}
where $c_i$, $\alpha_i$ represent the color and density computed from the $i$-th 3D Gaussian.

\begin{figure*}
  \centering
  \includegraphics[width=\textwidth]{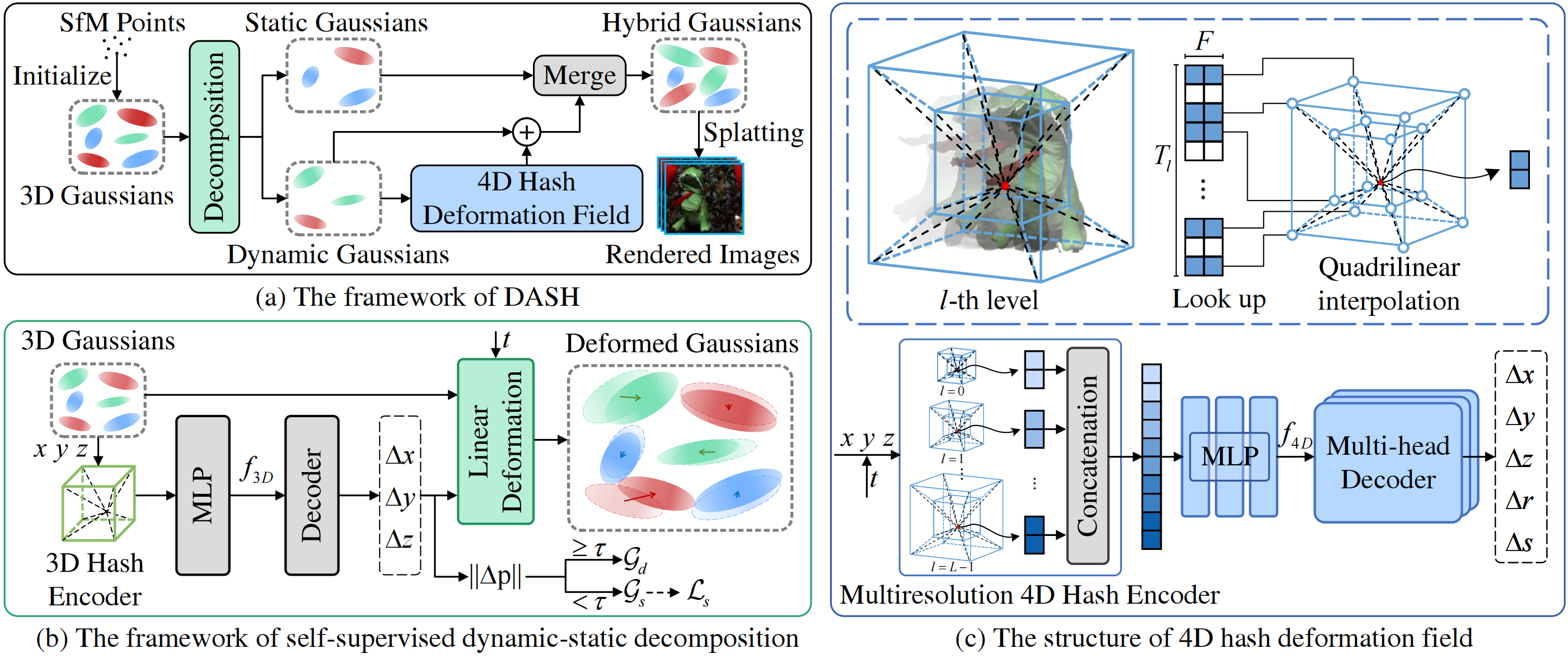}
  \caption{(a). The overview of \name, a real-time dynamic scene rendering framework that employs 4D hash encoding coupled with self-supervised decomposition. (b) Decomposition is guided by motion magnitude $||\Delta \textbf{p}||$ and optimized with our static constraint $\mathcal{L}_s$, where threshold $\tau$ is set at the top k\% percentile of $||\Delta \textbf{p}||$. (c) Our multiresolution 4D hash encoder takes dynamic Gaussian positions and timestamp as input to extract spatio-temporal features without relying on the low-rank assumption.}
  \label{fig:framework}
  \vspace{-3mm}
\end{figure*}

\section{Method}
\label{sec:method}

The overview of \name is illustrated in \cref{fig:framework} \textcolor{iccvblue}{(a)}.
Directly applying a 4D hash encoding to the entire scene requires a larger hash table due to increased hash collisions from the additional time dimension. To overcome this challenge, we primarily apply the 4D hash encoder to dynamic regions. We begin by decomposing the scene into static and dynamic components without additional supervision like manual annotations or precomputed masks, as shown in \cref{fig:framework} \textcolor{iccvblue}{(b)}.
After that, to effectively capture dynamic deformations, we present a 4D hash deformation filed. As shown in \cref{fig:framework} \textcolor{iccvblue}{(c)}, we propose a multiresolution 4D hash encoder that addresses the prevalent issues of low-rank assumption and feature overlap in current method. The detailed architecture is presented in \cref{sec:4dhash}. Finally, to enhance rendering fidelity and suppress chaotic artifacts, we introduce a spatio-temporal smoothness regularization in \cref{sec:optimization}.

\subsection{Self-Supervised Dynamic-Static Decomposition}
\label{sec:decompose}
In this section, we present our self-supervised dynamic-static decomposition method without  additional supervision. Critically, our method leverages the inherent behavioral difference between static and dynamic elements during optimization. As shown in \cref{fig:framework} \textcolor{iccvblue}{(b)}, we leverage this difference through a motion-aware feature extraction mechanism, combined with a simple yet effective linear model incorporating our proposed static constraint.

\paragraph{Motion-Aware Feature Extraction.}
In the original hash encoding~\cite{instant-ngp}, the isotropic sampling assumption holds in 3D space. However, in 4D space, the sampling is typically anisotropic~\cite{grid4d}. Therefore, when applying 3D hash encoding in 4D space to encode 3D coordinates, the spatial features can implicitly capture motion patterns due to the temporal variation in feature distributions. As shown in \cref{fig:framework} \textcolor{iccvblue}{(b)}, we extract spatial features $f_{3D}$ using a 3D hash encoder $G_{3D}$ from Instant-NGP~\cite{instant-ngp} followed by a tiny MLP $\phi_p$:
\begin{equation}
    f_{3D} = \phi_p(G_{3D}(\mathbf{p})).
    \label{eq:3dhash}
\end{equation}
Then MLP $\phi_l$ is employed to decode the features $f_p$ into positional deformation $\Delta\mathbf{p} = \phi_l(f_{3D})$.

\paragraph{Motion Modeling.}
We model the temporal trajectory of each Gaussian as linear motion:
\begin{equation}
    \mathbf{p}(t) = \mathbf{p} + t \Delta \mathbf{p} ,
    \label{eq:linear_motion}
\end{equation}
where $t$ is the time coordinate. Although real-world motion is often nonlinear, our simple linear approximation leads to two key observations: (1) for static Gaussians ${\mathcal{G}_s}$, the actual deformation satisfies $\Delta \mathbf{p}_s \equiv 0$, and (2) for dynamic Gaussians ${\mathcal{G}_d}$, the predicted $\Delta \mathbf{p}_d$ exhibits non-convergent oscillations during optimization, as linear motion cannot fully capture complex trajectories. These findings allow us to constrain $\Delta \mathbf{p}$ and interpret it as a first-order motion descriptor. The magnitude $||\Delta \mathbf{p}||$ directly corresponds to the Gaussian's average speed. This speed-based interpretation provides a physically grounded metric for motion intensity, enabling the effective decomposition of dynamic and static components without explicit supervision.

\paragraph{Self-Supervised Decomposition.}
We compute motion magnitudes $||\Delta \mathbf{p}||$ for all Gaussians and set a threshold $\tau$ at the top k\% percentile of these values. Gaussians with $||\Delta \mathbf{p}|| > \tau$ are classified as dynamic Gaussians ${\mathcal{G}_d}$, while the rest are classified as static Gaussians ${\mathcal{G}_s}$.
Static Gaussians are constrained by:
\begin{equation}
    \mathcal{L}_s =\frac{1}{N_s} \sum_{\mathbf{p} \in {\mathcal{G}_s}}||\Delta \mathbf{p}|| ,
    \label{eq:ls}
\end{equation}
where $N_s$ is the number of static Gaussians.
Our total loss function for decomposition follows the formulation in 3DGS~\cite{gaussian-splatting}, comprising a weighted combination of L1 color loss, D-SSIM loss, and our proposed static constraint:
\begin{equation}
    \mathcal{L}_d = (1 - \lambda_c) \mathcal{L}_1 + \lambda_c \mathcal{L}_{D-SSIM} + \lambda_s \mathcal{L}_s ,
\label{eq:decompose_total_loss}
\end{equation}
where $\lambda_c, \lambda_s$ are the hyperparameters to balance the losses.

To further enhance our dynamic scene representation, we employ the adaptive density control mechanism from 3DGS~\cite{gaussian-splatting}.
We apply this mechanism independently to both static and dynamic Gaussians, utilizing distinct configuration sets for each type.
This decoupled implementation allows ${\mathcal{G}_s}$ and ${\mathcal{G}_d}$ to optimize their spatial distributions according to their distinct motion characteristics. 


\subsection{4D Hash Encoding}
\label{sec:4dhash}
\paragraph{Multiresolution Hash Encoder.}
The original InstantNGP~\cite{instant-ngp} was designed primarily to render static 3D scenes. To address its limitations in modeling dynamic scenes, we propose a 4D multiresolution hash encoding that enables direct representation of dynamic Gaussians within a unified spacetime framework.
Following InstantNGP~\cite{instant-ngp}, we set the multiple resolutions of each dimension in a geometric progression between the coarsest and finest resolutions $[N_{min}, N_{max}]$:
\begin{equation}
    N_l = \lfloor N_{min} \cdot b \rfloor, \ b = \exp{\left(\frac{\ln{N_{max}} - \ln{N_{min}}}{L - 1}\right)} ,
    \label{eq:4dhash_dimension}
\end{equation}
where $l\in [0,L-1]$ indicates the level index, $L$ represents the total number of resolution levels, and $N_l$ is the resolution of level $l$.
The grid hypervoxel positions for the input $\mathbf{x}=(\mathbf{P}, t)$ could be calculated by rounding down and up in each level $\lfloor \mathbf{x}_l \rfloor = \lfloor \mathbf{x} \cdot N_l \rfloor, \lceil \mathbf{x}_l \rceil = \lceil \mathbf{x} \cdot N_l \rceil$.
The hypervoxels in each level could be obtained from the hash table by hashing the corresponding positions:
\begin{equation}
    h_l(\mathbf{x}_l) = \left( \bigoplus_{i=1, x_i \in \mathbf{x}_l}^d x_i \pi_i \right) \mod T_l ,
    \label{eq:4dhash_query}
\end{equation}
where $\bigoplus$ is the bit-wise XOR operation, $d$ is the dimension of the input, $\pi_i$ are unique large prime numbers, $T_l$ is the size of the level $l$ hash table, which stores learnable feature vectors with dimension $F$.
The encoded features can be computed through quadrilinear interpolation of the corresponding hypervoxel values in the 4D grid, followed by a tiny MLP $\phi_d$. Formally, given a 4D coordinate $\mathbf{x}=(\mathbf{P}, t)$, its encoded features from the 4D grid hash encoder $G_{4D}$ can be expressed as:
\begin{equation}
    f_{4D}=\phi_d(f_h),\ f_h=G_{4D}(\mathbf{x}) \in \mathbb{R}^{LF}.
    \label{eq:4dhash_feature}
\end{equation}

While 4D hash encoding provides efficient feature extraction, it suffers from $O(n^4)$ spatial complexity, leading to high collision rates and excessive storage for full-scene encoding~\cite{masked-spacetime-hashing}. Our decomposition method effectively mitigates these issues by focusing 4D hash encoding solely on dynamic Gaussians, thereby reducing storage and collisions while preserving fast query speeds.

\paragraph{Multi-head Deformation Decoder.}
Given the encoded features of dynamic Gaussians, we employ a multi-head deformation decoder $\mathcal{D}=\{\phi_x, \phi_r, \phi_s\}$ to predict the transformation parameters. The decoder consists of three dedicated MLP branches that respectively compute the deformation of position $\Delta \mathbf{x} = \phi_x(f_{4D})$, rotation $\Delta r= \phi_r(f_{4D})$, and scaling $\Delta s=\phi_s(f_{4D})$.
The deformed feature $(\mathbf{x}^\prime,r^\prime,s^\prime)$ can be addressed as:
\begin{equation}
    (\mathbf{x}^\prime, r^\prime, s^\prime) = (\mathbf{x} + \Delta \mathbf{x}, r + \Delta r, s + \Delta s).
    \label{eq:4dhash_deform}
\end{equation}
Then, we obtain the deformed dynamic Gaussians ${\mathcal{G}_d}^\prime=\{\mathbf{x}^\prime, s^\prime, r^\prime,\sigma,{C} \}$. Finally, we integrate the static Gaussians ${\mathcal{G}_s}$ with the deformed dynamic Gaussians ${\mathcal{G}_d}^\prime$ to construct the complete scene representation.

\subsection{Optimization with Smooth Regularization}
\label{sec:optimization}
Although the 4D hash encoder effectively captures spatio-temporal features, it lacks smoothness. This is a typical issue of explicit representation methods~\cite{grid4d}. Therefore, we set our regularization in the feature space. Generally, to regularize the 4D hash encoder, we propose a novel smooth regularization loss:
\begin{equation}
    \mathcal{L}_r = ||G_{4D}(\mathbf{x}) - G_{4D}(\mathbf{x} + \epsilon_\mathbf{x})||_2^2 ,
    \label{eq:smooth}
\end{equation}
where $\epsilon_\mathbf{x}=(\epsilon_{\mathbf{P}}, \epsilon_t)$ denotes a small random perturbation vector applied to the input coordinates $\mathbf{x}=(\mathbf{P}, t)$.
This regularization establishes spatio-temporal consistency constraints on the encoded features within local neighborhoods, guaranteeing smooth deformation patterns across neighboring Gaussians.
Our total loss function follows the formulation in 3DGS~\cite{gaussian-splatting}, comprising a weighted combination of L1 color loss, D-SSIM loss, and our proposed smooth regularization term:
\begin{equation}
    \mathcal{L} = (1 - \lambda_c) \mathcal{L}_1 + \lambda_c \mathcal{L}_{D-SSIM} + \lambda_r \mathcal{L}_r ,
\label{eq:total_loss}
\end{equation}
where $\lambda_c, \lambda_r$ are the hyperparameters to balance the losses.

\begin{table*}[tbp]
    \centering
    \begin{threeparttable}
        \tabcolsep=0.08cm
        \begin{tabular}{c|cc|cc|cc|cc|cc|cc|cc}
            \toprule
            \multirow{2}{*}{Method} & \multicolumn{2}{c}{Coffee Martini} & \multicolumn{2}{c}{Cook Spinach} & \multicolumn{2}{c}{Cut Beef} & \multicolumn{2}{c}{Flame Salmon$^1$} & \multicolumn{2}{c}{Flame Steak} & \multicolumn{2}{c}{Sear Steak} & \multicolumn{2}{c}{Mean} \\ 
            \cmidrule(lr){2-3}\cmidrule(lr){4-5}\cmidrule(lr){6-7}\cmidrule(lr){8-9}\cmidrule(lr){10-11}\cmidrule(lr){12-13}\cmidrule(lr){14-15}
            &PSNR&SSIM &PSNR&SSIM &PSNR&SSIM &PSNR&SSIM &PSNR&SSIM &PSNR&SSIM &PSNR&SSIM \\ 
            \midrule
            NeRFPalyer~\cite{nerfplayer} & \cellcolor{tablebest}31.53 &-&30.58&-&29.35&-&\cellcolor{tablebest}31.65&-&31.93&-&29.13&-&30.70&-\\
            HyperReel~\cite{hyperreel} &28.37&-&32.30&-&32.92&-&28.26&-&32.20&-&32.57&-&31.10&-\\
            K-Planes~\cite{k-planes}& \cellcolor{tablesecond}29.99 & 0.943 & 32.60 & 0.968 & 31.82 & 0.965 & \cellcolor{tablesecond}30.44 & 0.942 & 32.39 & 0.970 & 32.52 & 0.971 & 31.63 & 0.960 \\
            MixVoels~\cite{mixvoels}& \cellcolor{tablethird}29.36 & 0.946 & 31.65 & 0.965 & 31.30  & 0.965 & 29.92 & 0.945 & 31.21  & 0.970 & 31.43 & 0.971 & 30.81 & 0.960 \\
            E-D3DGS~\cite{ed3dgs} & 29.10 & \cellcolor{tablesecond}0.947 & \cellcolor{tablesecond}32.95 & 0.957 & \cellcolor{tablesecond}33.56  & 0.970 & 29.61 & \cellcolor{tablethird}0.949 & \cellcolor{tablesecond}33.57  & 0.974 & \cellcolor{tablethird}33.45 & 0.974 & \cellcolor{tablesecond}32.04 & 0.963 \\
            4DGaussian~\cite{4dgaussian} & 28.39 & 0.944 & 32.61 & \cellcolor{tablethird}0.971 & 32.07 & 0.966 & 29.14  & 0.948 & \cellcolor{tablethird}33.43 & \cellcolor{tablethird}0.977 & 32.85 & \cellcolor{tablethird}0.977 &31.42 & \cellcolor{tablethird}0.964\\
            Grid4D~\cite{grid4d} & 28.34 & 0.938 & 32.44 & \cellcolor{tablethird}0.971 & 33.23  & \cellcolor{tablethird}0.974 & 28.89 & 0.947 & 32.20  & \cellcolor{tablebest}0.980 & 33.15 & \cellcolor{tablesecond}0.978 & 31.38 & \cellcolor{tablethird}0.964 \\
            Grid4D+dec& 28.72 & \cellcolor{tablebest}0.950 & \cellcolor{tablethird}32.64 & \cellcolor{tablesecond}0.974 & \cellcolor{tablethird}33.51  & \cellcolor{tablebest}0.980 & 29.51 & \cellcolor{tablesecond}0.953 & 32.49  & 0.976 & \cellcolor{tablesecond}33.58 & \cellcolor{tablesecond}0.978 & \cellcolor{tablethird}31.74 & \cellcolor{tablesecond}0.967 \\
            Ours~\cite{grid4d}   & 28.92 & \cellcolor{tablesecond}0.947 & \cellcolor{tablebest}33.16 & \cellcolor{tablebest}0.980 & \cellcolor{tablebest}33.78  & \cellcolor{tablesecond}0.977 & \cellcolor{tablethird}29.94 & \cellcolor{tablebest}0.954 & \cellcolor{tablebest}33.77  & \cellcolor{tablesecond}0.979 & \cellcolor{tablebest}33.76 & \cellcolor{tablebest}0.979 & \cellcolor{tablebest}32.22 & \cellcolor{tablebest}0.969 \\
            \bottomrule
        \end{tabular}  
    \end{threeparttable}
    \caption{Quantitative comparisons on Neural 3D Video~\cite{dynerf} dataset. $^1$Using the first 300 frames of the 1200 frame long video. The \colorbox{tablebest}{best}, the \colorbox{tablesecond}{second best}, and the \colorbox{tablethird}{third best} are colored in table cells. Results of NeRF-based methods are from their original paper, while we calculate metrics of all Gaussian-based methods by running their official codes.}
    \label{tab:neu3d_comparison}
\end{table*}

\begin{figure*}
    \centering
    \includegraphics[width=\textwidth]{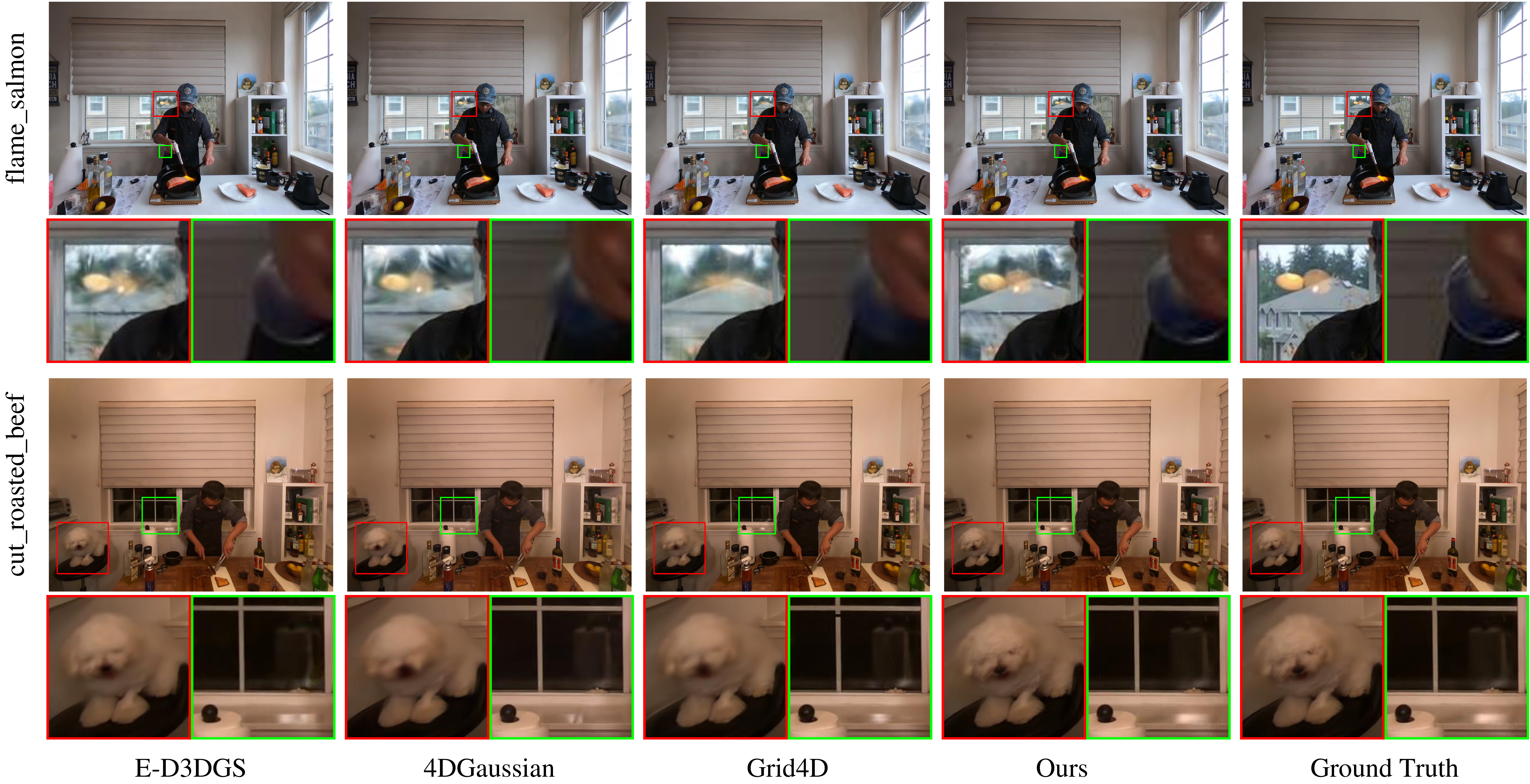}
    \caption{Qualitative comparisons on Neural 3D Video~\cite{dynerf} dataset.}
    \label{fig:neu3d_comparison}
    \vspace{-4mm}
\end{figure*}

\section{Experiment}
\label{sec:experiment}

In this section, we present a comprehensive experimental evaluation of our method across diverse real-world datasets. We compare with several state-of-the-art methods to validate the effectiveness of our approach. Furthermore, to further demonstrate the generality of our dynamic-static decomposition mechanism, we integrate it into Grid4D~\cite{grid4d} (denoted as Grid4D+dec) and conduct a thorough performance evaluation.

\subsection{Experimental Settings}
\label{sec:settings}

\paragraph{Datasets.}
We test on two real-world datasets.
Neural 3D Video~\cite{dynerf} dataset includes 15-20 multi-view videos with 300 frames each. Total six scenes are utilized to train and render at a resolution of 1352$\times$1014.
Technicolor Light Field~\cite{technicolor} dataset comprises synchronized 4$\times$4 camera array recordings with a resolution of 2048$\times$1088. Following HyperReel’s protocol~\cite{hyperreel}, we evaluate on five scenes at full resolution, using the center camera for testing, with 50 frames per scene.
For all datasets, our use COLMAP~\cite{colmap} point clouds from the first frame for initialization.

\begin{table*}[tbp]
    \centering
    \begin{threeparttable}
        \tabcolsep=0.40cm
        \begin{tabular}{c|ccc|c|c}
            \toprule
            Method  & PSNR & SSIM & LPIPS & Training (hours) $\downarrow$ & Storage (MB) $\downarrow$ \\
            \midrule
            DyNeRF~\cite{dynerf} & 31.80 & \cellcolor{tablesecond}0.958 & 0.142 & $>1000$ & \cellcolor{tablebest}30 \\
            HyperReel~\cite{hyperreel} & 32.73 & - & \cellcolor{tablethird}0.109 & \cellcolor{tablesecond}1.5 & \cellcolor{tablesecond}60 \\
            E-D3DGS~\cite{ed3dgs} & \cellcolor{tablesecond}33.16 & \cellcolor{tablethird}0.955 & \cellcolor{tablesecond}0.105 & 2.93 & \cellcolor{tablethird}76 \\
            Grid4D~\cite{grid4d} & 32.59 & 0.949 & 0.128 & \cellcolor{tablebest}1.2 & 319 \\
            Grid4D+dec & \cellcolor{tablethird}33.04 & \cellcolor{tablethird}0.955 & 0.114 & \cellcolor{tablethird}2 & 192 \\
            Ours & \cellcolor{tablebest}33.94 & \cellcolor{tablebest}0.960 & \cellcolor{tablebest}0.097 & \cellcolor{tablesecond}1.5 & 146 \\
            \bottomrule
        \end{tabular}
    \end{threeparttable}
    \vspace{-1mm}
    \caption{Quantitative comparisons on Technicolor Light Field~\cite{technicolor} dataset. The \colorbox{tablebest}{best}, the \colorbox{tablesecond}{second best}, and the \colorbox{tablethird}{third best} are colored in table cells. Results of NeRF-based methods are from their original paper, while we calculate metrics of all Gaussian-based methods by running their official codes. The training time and storage data for DyNeRF~\cite{dynerf} and HyperReel~\cite{hyperreel} are estimated based on the official implementations, while we calculate these metrics of all Gaussian-based methods in the Painter scene.}
    \label{tab:technicolor_comparison}
\end{table*}

\paragraph{Implementation Details.}
Our implementation is primarily based on the PyTorch~\cite{pytorch} framework, with the 4D hash encoder implemented using CUDA/C++. All experiments were conducted on a single NVIDIA RTX 4090 GPU. We’ve fine-tuned our optimization parameters by the configuration outlined in the 3DGS~\cite{gaussian-splatting}. We evaluate the quality of rendered images using peak-signal-to-noise ratio (PSNR), structural similarity index (SSIM~\cite{ssim}), and learned perceptual image patch similarity (LPIPS~\cite{lpips}). 

\paragraph{Baselines.}
We compare \name with several state-of-the-art models~\cite{nerfplayer,mixvoels, hyperreel, dynerf, ed3dgs, 4dgaussian, grid4d}. NeRFPlayer~\cite{nerfplayer}, K-Planes~\cite{k-planes}, MixVoels~\cite{mixvoels}, DyNeRF~\cite{dynerf} and HyperReel~\cite{hyperreel} are NeRF-based dynamic rendering approaches.
4DGaussian~\cite{4dgaussian}, E-D3DGS~\cite{ed3dgs} and Grid4D~\cite{grid4d} are Gaussian-based dynamic rendering methods.

\begin{figure}
    \centering
    \includegraphics[width=8.3cm]{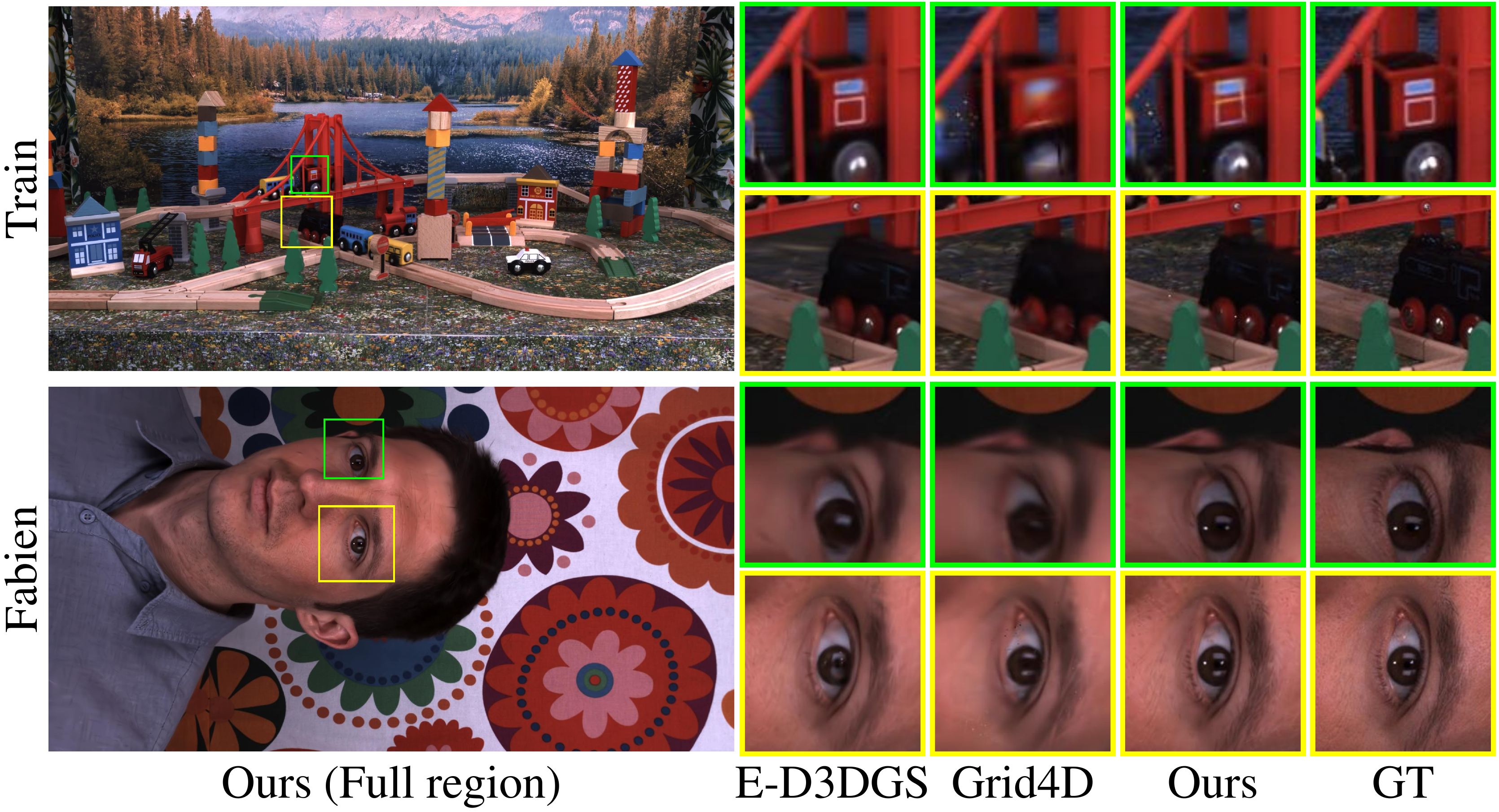}
    \caption{Qualitative comparisons on the Technicolor Light Field~\cite{technicolor} dataset. GT denotes the ground truth.}
    \label{fig:technicolor_comparison}
\end{figure}

\begin{figure}
    \centering
    \includegraphics[width=8.3cm]{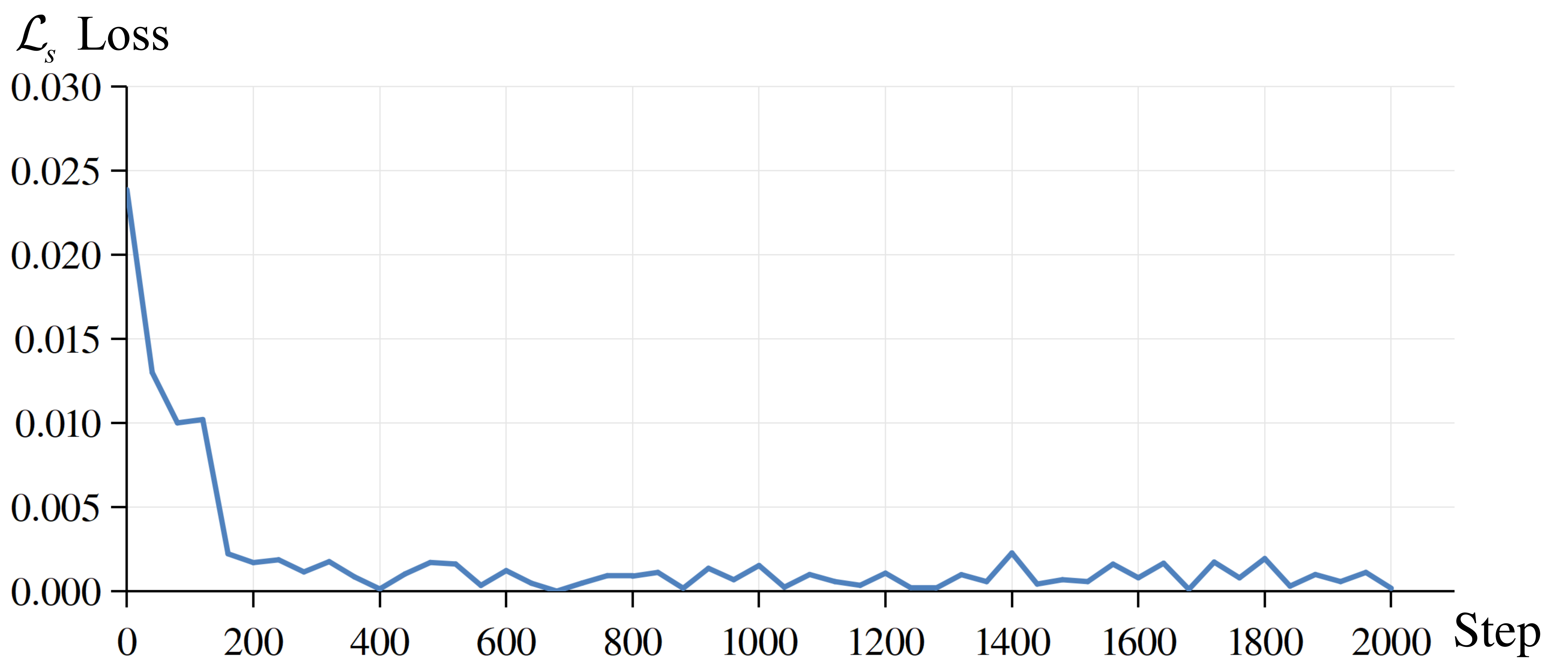}
    \caption{Convergence of the loss $\mathcal{L}_s$ in Painter scene from Technicolor Light Field~\cite{technicolor} dataset.}
    \label{fig:ls}
    \vspace{-3mm}
\end{figure}

\subsection{Results and Comparisons}
\paragraph{Results on Neural 3D Video.}
\cref{tab:neu3d_comparison} presents per-scene results on the Neural 3D Video dataset. 4DGaussians~\cite{4dgaussian} suffers from feature overlap and detail loss due to its low-rank limitation. In contrast, our method resolves these issues, achieving state-of-the-art performance with a mean PSNR of 32.22 dB.
Qualitative comparisons in \cref{fig:neu3d_comparison} substantiate our method's superiority, with enhancements in detail preservation and artifact mitigation in red and green boxes. Our \name excels in rendering complex details like the kitchen torch and white dog. Besides, \cref{fig:neu3d_comparison} also shows superior performance in rendering specular reflections and transmissions relative to competing approaches.

\begin{figure*}
  \centering
  \includegraphics[width=\textwidth]{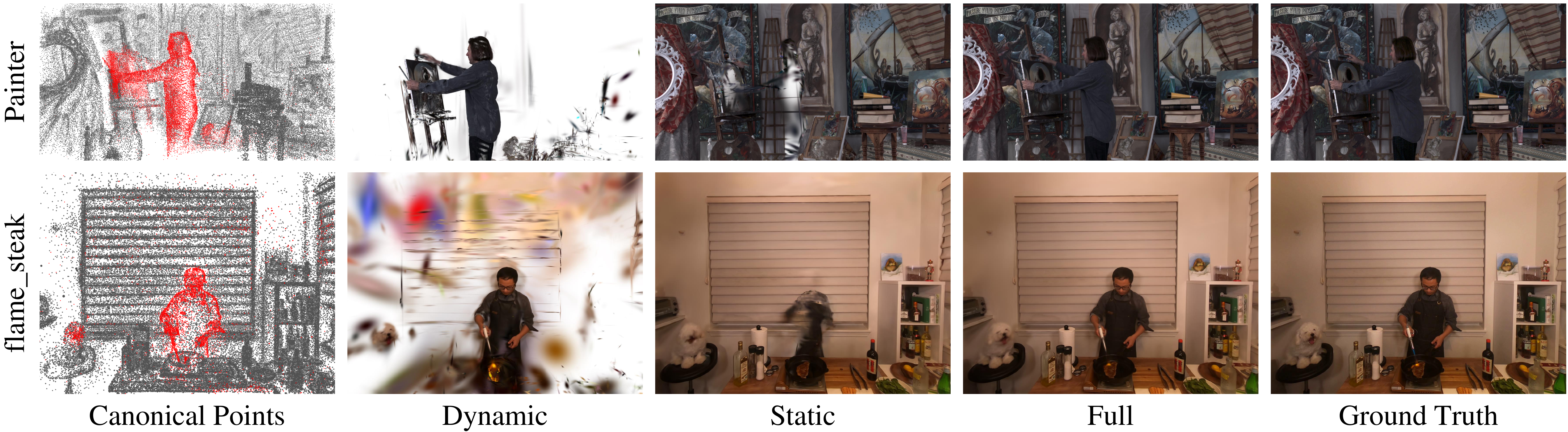}
  \caption{Visualization of our dynamic-static decomposition on Neural 3D Video~\cite{dynerf} dataset and Technicolor Light Field~\cite{technicolor} dataset. In the canonical points, dynamic components are marked in red and static components are marked in black.}
  \label{fig:visualization_decomposition}
  \vspace{-2mm}
\end{figure*}

\paragraph{Results on Technicolor Light Field.}
\cref{tab:technicolor_comparison} presents quantitative results on the Technicolor Light Field dataset.
Our method achieves a state-of-the-art 33.94 dB PSNR that significantly outperforms existing methods.
\cref{fig:technicolor_comparison} illustrates qualitative comparisons.  Grid4D exhibits blurred reconstructions and detail loss, attributable to feature overlap, resulting in the visibly blurred train and over-smoothed eyes. E-D3DGS, while offering some improvement with its MLP deformation fields, still produces over-smoothed details, particularly noticeable in the background of the black train and the eyes. In contrast, our approach effectively addresses these shortcomings, producing sharper outlines and preserving fine structural details. More results are available in the supplementary material.

\paragraph{Comparisons of FPS and Training Time.}
Comparing Frames Per Second (FPS) directly can be misleading, as it is significantly influenced by the varying number of Gaussians employed by different models. Consequently, we present both FPS and the corresponding Gaussian counts in \cref{tab:fps} to provide a more fair performance evaluation. It is evident that NeRF-based methods commonly suffer from long training times and slow rendering speeds. In contrast to Gaussian-based approaches, our method achieves enhanced rendering performance while leveraging a reduced number of Gaussians. This efficiency is attributed to the $O(1)$ query complexity of hash table and the strategic focus on modeling dynamic components only.

\begin{table}
    \centering
    \begin{threeparttable}
        \tabcolsep=0.16cm
        \begin{tabular}{c|ccc}
            \toprule
            Method & {N $\downarrow$} & {Training (hours) $\downarrow$} & {FPS $\uparrow$} \\
            \midrule
            NeRFPlayer~\cite{nerfplayer} &-&6&0.045\\
            HyperReel~\cite{hyperreel}&-&9&2.0\\
            K-Planes~\cite{k-planes}&-&1.8&0.3\\
            MixVoels~\cite{mixvoels}&-&\textbf{0.3}&15\\
            E-D3DGS~\cite{ed3dgs} & 263.1k & 4.9 & 39.3 \\
            4DGaussian~\cite{4dgaussian} & 123.7k & 0.8 & 136.9 \\
            Grid4D~\cite{grid4d} & 158.9k & 1 & 256.8 \\
            Ours & \textbf{106.6k} & 1 & \textbf{264.3} \\
            \bottomrule
        \end{tabular}
    \end{threeparttable}
    \caption{Comparisons of Gaussian number, training time and FPS in the coffee martini scene from Neural 3D Video~\cite{dynerf} dataset. The NeRF-based results originate from the 4DGaussian~\cite{4dgaussian} paper.}
    \label{tab:fps}
\end{table}

\begin{figure}
    \centering
    \includegraphics[width=8.3cm]{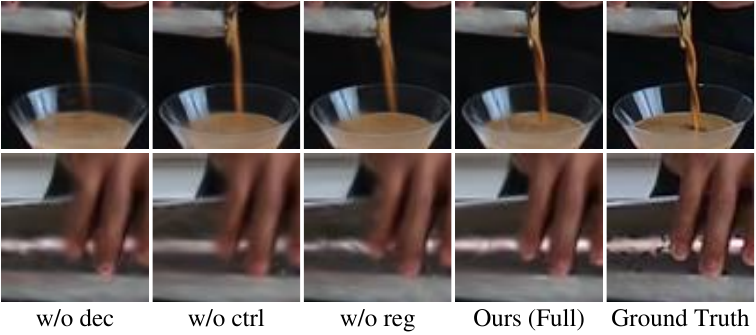}
    \caption{Qualitative results of our ablation studies.}
    \label{fig:ablation}
    \vspace{-3mm}
\end{figure}

\paragraph{Visualization of Dynamic-Static Decomposition.}
We present the convergence curve of our static constraint loss $\mathcal{L}_s$ during the decomposition training process. As shown in \cref{fig:ls}, $\mathcal{L}_s$ converges rapidly, ensuring stable optimization and accurate identification of static components.
Moreover, our comprehensive visualizations in \cref{fig:visualization_decomposition} present both dynamic and static Gaussian renderings with their canonical space distributions, demonstrating the effectiveness of our self-supervised decomposition method. Furthermore, when implemented in Grid4D (referred to as Grid4D+dec), the method achieves substantial PSNR gains, as quantitatively verified in \cref{tab:neu3d_comparison} and \cref{tab:technicolor_comparison}. These consistent qualitative and quantitative results confirm our method's capability in scene decomposition. 

\subsection{Ablation Studies}
\paragraph{Ablation of Dynamic-Static Decomposition.}
To validate the necessity of dynamic-static decomposition, we conduct an ablation study by applying 4D hash encoder to the entire scene without decomposition (referred to as w/o dec). \cref{tab:ablation} and \cref{fig:ablation} show that this approach suffers from performance degradation due to high hash collision rates in dense points, which particularly impair dynamic region representation. By isolating static components, our decomposition method effectively mitigates hash conflicts, enhancing deformation field accuracy and overall performance.

\paragraph{Ablation of Dynamic Density Control.}
We implement independent adaptive density control and opacity reset mechanisms for static and dynamic Gaussians, following 3DGS~\cite{gaussian-splatting}. As shown in \cref{tab:ablation} and \cref{fig:ablation}, omitting density control for dynamic Gaussians (referred to as w/o ctrl) severely impacts the rendering quality, 
since the initial decomposition only provides a preliminary distribution of dynamic regions, which still requires subsequent refinement to achieve a more accurate representation of details.
\paragraph{Ablation of Smooth Regularization.}
The proposed smooth regularization aims to mitigate the chaotic deformation prediction. We train \name without the smooth regularization as w/o reg. \cref{tab:ablation} and \cref{fig:ablation} show that the regularization establishes spatio-temporal consistency constraints within local neighborhoods, guaranteeing smooth deformation patterns across neighboring Gaussians.

\begin{table}
    \centering
    \begin{threeparttable}
        \tabcolsep=0.5cm
        \begin{tabular}{c|ccc}
            \toprule
            Method & PSNR & SSIM & LPIPS\\
            \midrule
            w/o dec & 31.65 & 0.967 & 0.052 \\
            w/o ctrl & 31.92 & 0.967 & 0.054\\
            w/o reg & 31.51 & 0.966 & 0.056\\
            Full & \textbf{32.22} &  \textbf{0.969} & \textbf{0.050}\\
            \bottomrule
        \end{tabular}
    \end{threeparttable}
    \caption{Quantitative ablation results on Neural 3D Video~\cite{dynerf} dataset.}
    \label{tab:ablation}
    \vspace{-4mm}
\end{table}

\section{Conclusion}
\label{sec:conclusion}
In this paper, we propose \name, a novel framework for real-time dynamic scene rendering. \name introduces a self-supervised dynamic-static decomposition mechanism that effectively separates scene components without explicit annotations. In addition, we propose a multiresolution 4D hash encoder that addresses the prevalent issues of low-rank assumption and feature overlap in current methods. Furthermore, our spatio-temporal smoothness regularization improves the rendering quality. Extensive experiments demonstrate that \name achieves state-of-the-art performance while maintaining real-time rendering speeds.

\section{Acknowledgment}
\label{sec:acknow}
\begin{flushleft}
This work was in part supported by the National Natural Science Foundation of China under grants 62032006 and 62021001, and Open Fund of APKL of BIIP, IAI, Hefei Comprehensive National Science Center under grants 24YGXT003.
\end{flushleft}
{
    \small
    \bibliographystyle{ieeenat_fullname}
    \bibliography{main}
}


\end{document}